\documentclass[conference]{IEEEtran}
\IEEEoverridecommandlockouts
\usepackage{gensymb}
\usepackage{caption} 
\usepackage{enumitem}

\usepackage{dblfloatfix}

\usepackage{cite}
\usepackage{amsmath,amssymb,amsfonts}
\usepackage{graphicx}
\usepackage{textcomp}
\usepackage{xcolor}

\def\BibTeX{{\rm B\kern-.05em{\sc i\kern-.025em b}\kern-.08em
    T\kern-.1667em\lower.7ex\hbox{E}\kern-.125emX}}

\begin{document}

\title{A Temporal Convolution Network Approach to State-of-Charge Estimation in Li-ion Batteries}

\author{\IEEEauthorblockN{Aniruddh Herle$^{\gamma}$, \textit{Student Member},  Janamejaya Channegowda$^\delta$, \textit{Member, IEEE}, Dinakar Prabhu$^{\beta}$, \textit{Member, IEEE} \\
$^{\gamma}$Ramaiah Institute of Technology, Bangalore, India, aniruddh.herle@gmail.com \\
$^\delta$Ramaiah Institute of Technology, Bangalore, bcjanmay.edu@gmail.com}
$^\beta$Indian Institute of Science, Bangalore, dinakar0409@gmail.com}

\maketitle

\begin{abstract}
Electric Vehicle (EV) fleets have dramatically expanded over the past several years. There has been  significant increase in interest to electrify all modes of transportation. EVs are primarily powered by Energy Storage Systems such as Lithium-ion Battery packs. Total battery pack capacity translates to the available range in an EV. State of Charge (SOC) is the ratio of available battery capacity to total capacity and is expressed in percentages. It is crucial to accurately estimate SOC to determine the available range in an EV while it is in use. In this paper, a Temporal Convolution Network (TCN) approach is taken to estimate SOC. This is the first implementation of TCNs for the SOC estimation task. Estimation is carried out on various drive cycles such as  HWFET, LA92, UDDS and US06 drive cycles at 1 C and 25 \degree Celsius. It was found that TCN architecture achieved an accuracy of 99.1\%.\\
\end{abstract}


\section{Introduction}
\indent In order to meet international emission targets, a transition away from fossil-fuel based transportation is required \cite{Tong2019}. The transportation sector contributes to 25\% of the total Green House Gas (GHG) emissions of the world, automobiles being the primary source \cite{Khurana2020}. With rising global temperatures, this sector is moving towards the environmentally friendly Lithium Ion Battery powered Electric Vehicle (EV).\\ 
\indent The successful adoption of Electric Vehicle Technology is highly reliant on the operation of the Battery Management System (BMS). The role of the BMS is to collect real-time data from the Li-ion Battery (LIB) pack and use it to monitor the battery operation. Battery monitoring is crucial to the operation of the EV as well as the safety of the passengers. Therefore, the reliability of the BMS is of paramount importance to the safe operation of the EV \cite{Xu2018a}. The BMS collects cell parameters like voltage, current, temperature, State of Charge (SOC), State of Health (SOH), State of Power (SOP), and so forth \cite{How2019}.\\
\indent One of the key parameters to be monitored is SOC. The mathematical formula for SOC is given below:

\begin{equation}
SOC = \frac{Q_{available}}{Q_{rated}}
\end{equation}
Where $Q_{available}$ is the charge left in the battery and $Q_{rated}$ is the maximum charge the battery can hold.\\ \\

\indent The SOC value is crucial for the operation of the BMS. The importance of SOC estimation for the BMS is listed below:

\begin{enumerate}[label=(\roman*)]

\item SOC is used for several downstream calculations for cell balancing and SOH estimation
\item Accurate SOC values can be used to ensure battery pack operation within the limits specific to the battery
\item A BMS with accurate SOC values can optimize the cell usage within the battery pack, thereby increasing battery pack performance

\end{enumerate}

\begin{figure*}[b!]
    \centering
    \includegraphics[width=0.7\textwidth]{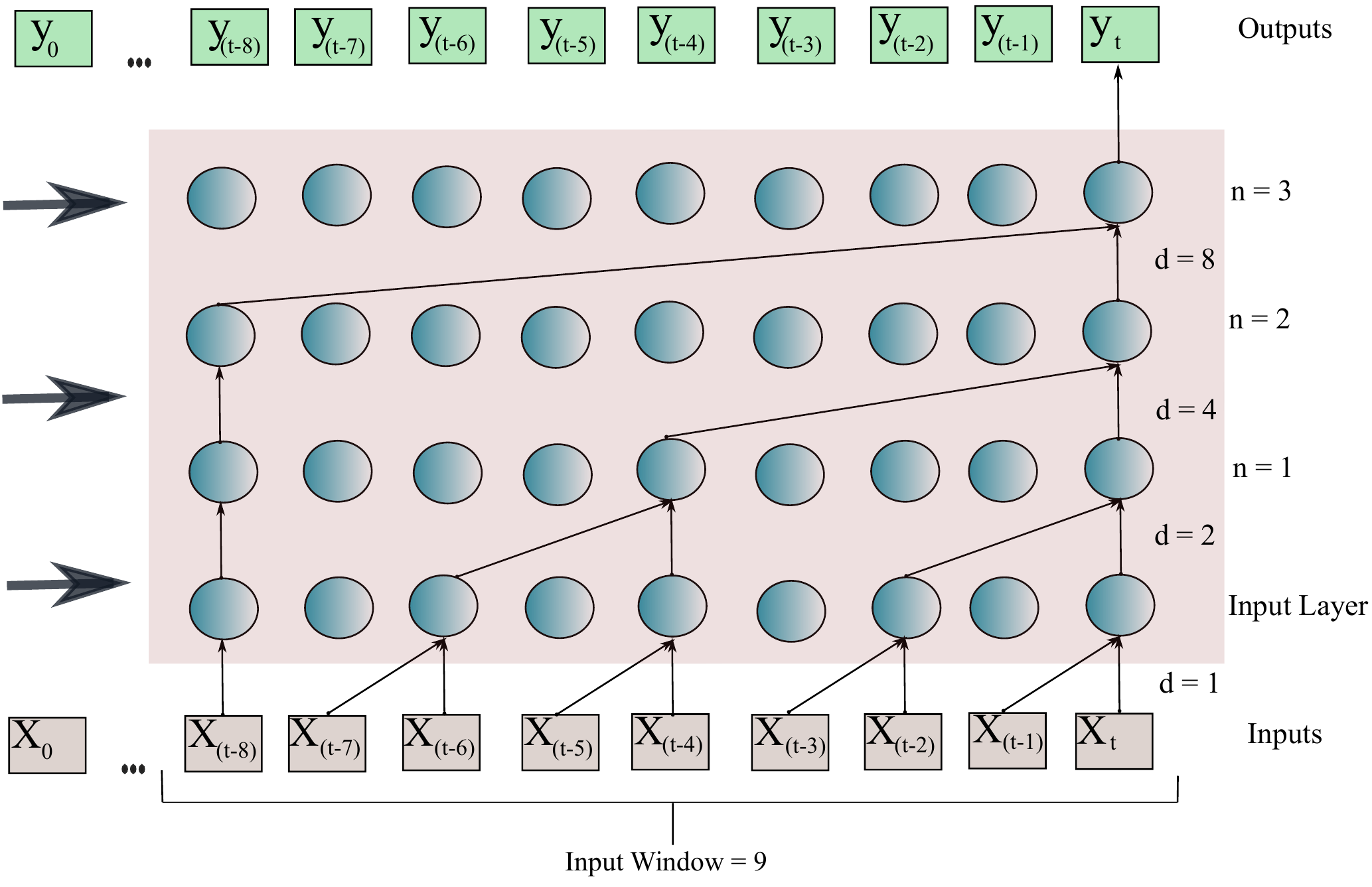}
    \caption{A single stack of the TCN model on a univariate system}
    \label{fig:1}
\end{figure*}
\indent SOC estimation, however, is not a straightforward task. Complex ageing and thermal effects come into play during EV operation, and the total battery capacity fades over time \cite{Cheng2011}. These factors contribute to making the cell dynamics highly non-linear. Moreover, SOC cannot be measured directly, and therefore must be estimated based on closely related proxy measures like voltage, current and temperature.\\
\indent Several traditional methods have been proposed to solve this problem, like Coulomb counting, Open Circuit Voltage, Electrochemical Impedance Spectroscopy, EMF Method and Model Based methods. The problem with these methods, however, is that they are highly error-prone. Developments in algorithmic approaches to measure SOC indirectly lead to advances in research in several filters like Particle Filter \cite{Schwunk2013}, H-infinity filter \cite{Yu2017} and several variations of the Kalman Filter \cite{Hussein2011, Zhang2008a, Sun2011}.\\ 
\indent Advances in computational resources in recent years have opened up opportunities for Machine Learning (ML) methods to be employed for the SOC estimation task. A large variety of ML algorithms have been used for this, like Support Vector Regression \cite{Anton2013} and Gaussian Process Regression \cite{Chehade2020}. State-of-the-art techniques like Deep Learning\cite{Shen2019, Shen2020} have also become more popular for SOC Prediction.\\
\indent Recurrent Neural Networks (RNNs) have been the mainstay of sequence modelling across many fields, and Li-ion battery technology is no exception. Non-linear Autoregressive Exogenous Input Neural Network (NARXNN) and Long-Short Term Memory (LSTM)-RNNs have become very popular. In Chemali et al. \cite{Chemali2018t, Chemali2018}, an LSTM-RNN was implemented and achieved an MAE between 0.69\% to 1.3\% in the temperature range 0 to 25 \degree Celsius. Khalid et al. \cite{Khalid2019} showed that optimizers can be used to reduce the MSE of LSTM-RNN to 0.22\%. In \cite{Lipu2018}, NARX was used together with Particle Swarm Optimisation (PSO) algorithm for the estimation of SOC resulting in an MAE from 0.41\% to 0.64\% in the temperature range 0 to 45 \degree C. In Chaungxin et al. \cite{Chuangxin2019} NARX was paired with Genetic Algorithm and achieved an MAE of 0.22 \%. Comparisons between NARXNN and LSTM have also been made in Abbas et al. \cite{Abbas2019}.\\
\indent The key drawback of RNN approaches, especially LSTM, is that they are computationally expensive. This is not a concern when training the model, but when deployed in Battery Management Systems in real-time, can pose problems because of the large memory space requirement. A key challenge for LIB technology is to develop accurate and robust models with low computational complexity that can be deployed even on memory-constrained hardware of the BMS \cite{Sidhu2019}. A fast and reliable SOC estimation model is required to keep the BMS updated in as close to real-time as possible. High model complexity can be detrimental in real on-vehicle applications, because the memory space and computation power required is high. A lower complexity model comes with the obvious drawback of lower accuracy. Essentially, researchers are working against the trade-off between Complexity and Accuracy.\\
\indent A novel machine learning algorithm called Temporal Convolution Network (TCN) is becoming increasingly popular in RNN dominated fields like audio-signal processing \cite{Jeong2016, Oord2016} and satellite image processing \cite{Pelletier2019}. TCNs have been shown to have superior performance compared to RNNs, which are usually the first choice. In Jayasinghe et al. \cite{Jayasinghe2019}, a TCN was used for Remaining Useful Life (RUL) prediction and more recently in Zhou et al. \cite{Zhou2020} for SOH prediction. To the best of our knowledge, this is the first work dedicated to the application of TCNs for the SOC estimation task.\\
\indent One of the improvements introduced by the LSTM-RNN framework is that they reduce the gradient explosion/vanishing problem. This occurs in deep recurrent neural networks when the gradient undergoes repeated matrix multiplications as it is propagated backwards to earlier layers. If the gradient value is small, it rapidly approaches zero, and if it is large, it rapidly reaches infinity. In the LSTM framework, this is mitigated by introducing a Forget Gate at each cell. Another important feature of LSTMs is that they produce an output of the same length as the input. The TCN architecture inherently possess both these traits, as well being much more efficient computationally. This work proposes the TCN architecture for the SOC estimation task, and an analysis of network hyper-parameter tuning is done to understand the feasibility of this approach for accurate SOC estimation. 

\section{Temporal Convolution Network}
\indent Temporal Convolution Networks (TCN) are a breed of neural networks that are becoming increasingly popular for sequence modelling tasks. TCNs are different from a conventional CNN with respect to the direction in which the convolution filter is applied to the inputs. Whereas a CNN works by applying a convolution spectrally, the TCN applies the convolution across different time steps of the input. This means that the convolution attempts to capture temporal variations in the data.\\
\indent These convolutions are causal, which means that the output of this type of convolution layer is dependent on the past and present values of the input, and is completely independent of the future values of the input. This indicates that the prediction at time step $t$ does not depend on future values of the input (values of input at times $t+1,t+2 ...$)\cite{Oord2016}. The problem with this is that a network comprised of causal convolutions can look back at a history size that scales only linearly with the depth of the network. This can be an issue for certain sequence modelling tasks which require long histories to be considered. To solve this, the convolutions are also dilated.\\
\indent To illustrate this, a simple univariate system is considered in Fig. \ref{fig:1}. As can be seen, a sliding window approach is used. In the example, a window length of 9 is used, and the Stack moves across the input sequence. The first layer of convolutions is applied to all the input features within the input window. This means that the dilation factor, $d$, is 1. All subsequent layers are applied on the outputs of the previous layers, like a typical CNN. This means that the convolution filter response (activation map \cite{Pelletier2019}) of the first layer is the input to the second layer. However, in the second layer, the convolution is applied across every second input, so that $d$ is 2. Similarly, the third layer looks at every fourth input, and the fourth layer every eighth. Thus, the dilation factor is made to scale exponentially with network depth. As a result, the number of time steps into the past that can be considered, or the look-back time, is increased significantly. The size of the receptive field of a TCN depends on the network depth (number of layers, $n$), filter size ($k$) and the dilation factor, $d$. Increasing any of these parameters increases the look-back time of the model. \\
\begin{figure}[t!]
    \centering
    \includegraphics[width=0.25\textwidth]{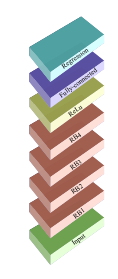}
    \caption{Layers of a double stack network of the TCN architecture}
    \label{fig:2}
\end{figure}
\indent The dilation factor is increased as powers of 2 in order to prevent what is called the ‘gridding effect’ \cite{Zhou2020}. This occurs when large swathes of data are missed out as the network gets deeper, which can be catastrophic for sequence modelling tasks. As only some regions get sampled by the higher layers, gaps are created in between the time series sequence instances \cite{Wang2018}. Restricting the value of $d$ to incremental powers of two ensures that every time instance is convolved, while at the same time ensuring that the model possesses a very large receptive size \cite{Bai2018}.\\
\indent The layers of each stack are shown in Fig. \ref{fig:2}. Each stack contains four Residual Blocks (RB), which each in turn is made up of different layers. The figure illustrates a two stack network. Each Residual Block is made up of two Convolution Layers, each with 132 total trainable parameters. There is a Dropout and ReLu operation for each Convolution Layer. A Regression Layer is used to give a final numerical output at each level in the network.\\
\indent Unlike RNNs, where previous time steps are processed only after the subsequent time step (sequentially) \cite{Benitez2020}, the convolution filters in a TCN can be applied in parallel across all time steps in the input window. This significantly speeds up both the training and prediction time \cite{Bai2018}. Moreover, the trained filter weights are shared across a layer, which means that the memory requirement is significantly lower for TCNs as compared to more popular RNNs like LSTM and GRU. In \cite{How2019}, one of the key issues to be solved in the SOC estimation task is the management of the accuracy-memory space trade off. The model must be accurate and robust, while still maintaining a low computational complexity. This is because the model should be deployable on low-cost BMS systems in which processing power and memory is constrained, and also be able to provide SOC estimations in real-time to the BMS.  As the same set of filter weights is used in different parts of the time series data, the number of trainable parameters in the network reduces significantly \cite{Pelletier2019}. Moreover, TCNs have been shown to outperform LSTM-RNN on several baseline sequence modelling tasks, which are typically considered to be the domain of this recurrent network \cite{Bai2018}.
A summary of the key advantages of TCN over other time series modelling algorithms is given below:

\begin{enumerate}[label=(\roman*)]

\item Immune to the gradient vanishing/ exploding problem
\item Output length is the same as input length, which is useful for certain applications
\item Large receptive field, due to the exponential increase of $d$
\item Convolution filters are applied in parallel across the input
\item Relatively low memory space requirement
\item Reduced computational complexity results in faster prediction times

\begin{figure*}[htbp!]
    \centering
    \includegraphics[width=0.7\textwidth]{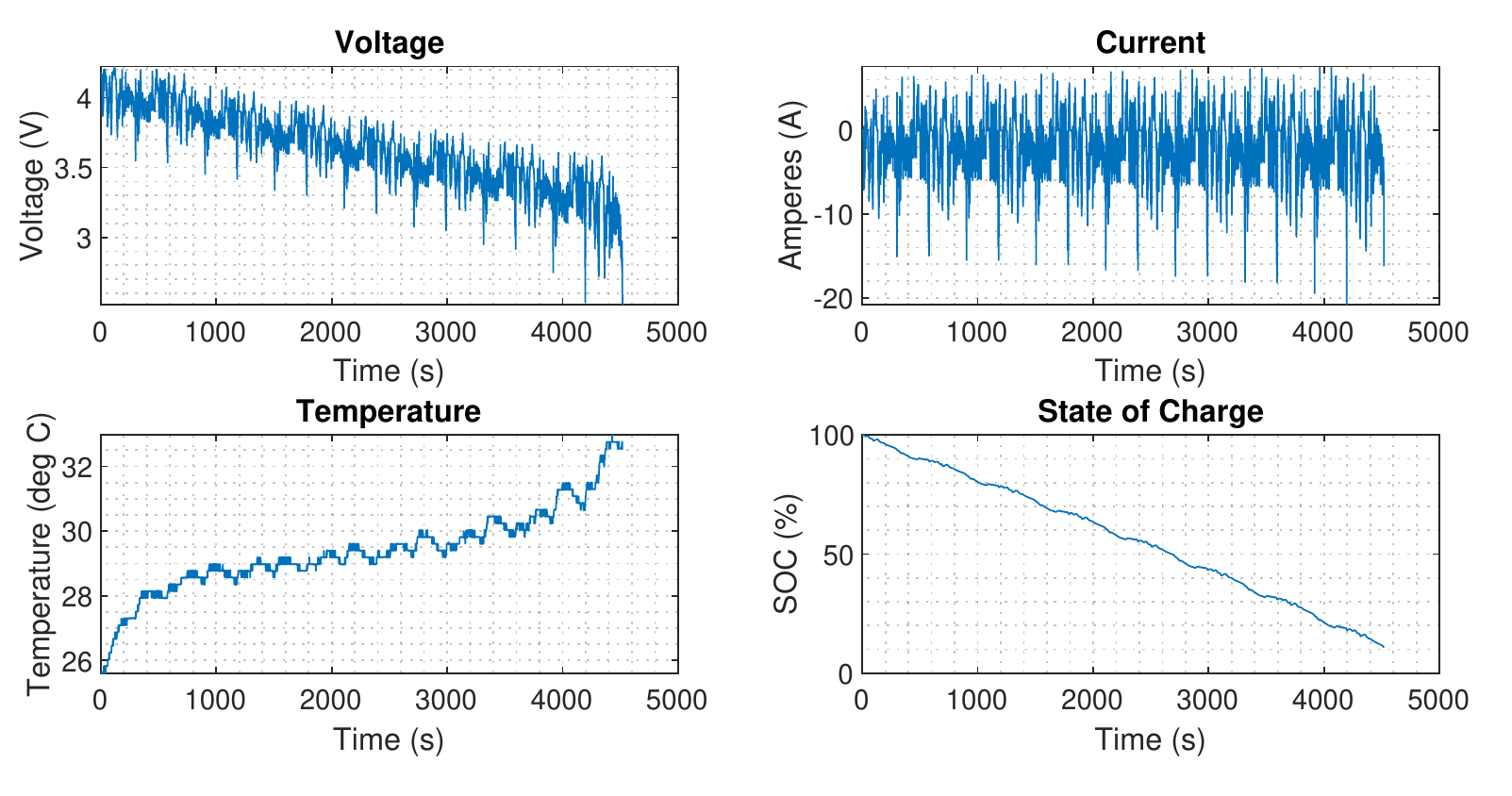}
    \caption{Voltage, Current, Temperature and SOC variation at 25 \degree C of the Panasonic 18650 cell}
    \label{fig:3}
\end{figure*}

\end{enumerate}

\section{Dataset}
\indent The dataset in \cite{Chemali2018} was used for this work. A Panasonic 18650 battery cell with a lithium nickel cobalt aluminium oxide (NCA) chemistry was subjected to different drive cycles at different temperatures (0 \degree C to 25 \degree C). The battery specifications are shown in Table \ref{table:2} and the variation of cell parameters is shown in Fig \ref{fig:3}.\\ 
\indent The battery testing equipment designed by Digatron Firing Circuits was used to acquire the training and validation data \cite{Chemali2018}. For this work, the 25 \degree Celsius case is considered. At this temperature, the battery specifications allow charging of the battery, which is only safe at temperatures greater than 10 \degree C \cite{Chemali2018}, so regenerative braking can also be factored into the training data. The current sensor used for the data acquisition has an error less than 25 mA, summing up to less than 40 mAh for the entire dataset. This is only 1.38\% of the battery capacity.\\
\indent The subset of the dataset used in this work considers data collected when the battery was exposed to 4 drive cycles, namely HWFET, LA92, US06 and UDDS. The drive cycles for a Ford F150 truck were used\cite{Kollmeyer2015}. A hybrid dataset made from a combination of all these drive cycles was used to train the model, as it provides an excellent real-life approximation. This allows the model to train on a broad set of battery dynamics.

\begin{table}[!htpb]
\begin{center}
\caption{Panasonic 18650PF Battery Specifications} 
\label{table:2}
\begin{tabular}{ |c|c| } 
 \hline
 Nominal Capacity & 2.9 Ah \\
 \hline
 Nominal Voltage & 3.6 V \\
\hline 
 Minimum Voltage & 2.5 V \\
\hline 
 Maximum Voltage & 4.2 V \\
\hline
 Power Density & 206.25 Wh/kg\\
 \hline
\end{tabular}
\end{center}
\end{table}

\section{Methodology}
\indent In order to find the optimal network hyper-parameters like number of stacks, input window and so on, several different models were trained and tested on four drive cycles. Python 3 was used to design the algorithm. Tensorflow was implemented using Keras as a frontend. A Tesla K80 GPU provided by Google Colaboratory was used to train the models. All the prediction times were calculated for this processor. The HWFET, LA92, UDDS and US06 drive cycles were used in this work. Together, they represent a close to real-life situation of the load profile the battery may be subjected to in practice. In order to train the model, a hybrid dataset made from a combination of these four drive cycles was used. Load profiles collected at 25 \degree C ambient temperature and 1 C was used. \\
\indent The inputs to the model were battery current, battery voltage, battery temperature and the past SOC values of the time steps within the range specified by the input window. The voltage, current and temperature values first underwent min-max normalisation, which aids in the convergence of deeper neural networks \cite{Benitez2020}.\\
\indent Three different input window settings were tested, while varying the number of stacks from 2 to 25. The input window lengths considered are 100, 500 and 1000 time steps. The trained models are then tested on HWFET, LA92, UDDS and US06 drive cycles and the accuracies and prediction time per data point (or time instance) is calculated. The average of these values across the four drive cycles are shown in Table \ref{table:3}.

\section{Results}

\begin{table*}[t]
\begin{center}
\caption{Variation of Memory Requirement, Accuracy and Prediction Time for different TCN architectures}
\label{table:3}
\begin{tabular}{|c|c|c|c|c|c|c|}
\hline
No. Stacks & Input Window & Parameters & MSE &  File Size (kB) & Average Prediction Time(ms) & Average Accuracy (\%)\\
\hline
         2 &        100 &       2033 &       0.01 &        733 &      0.051 &      96.27 \\

         4 &        100 &       4145 &     0.3131 &       1348 &      0.072 &      81.37 \\

         8 &        100 &       8369 &     0.0447 &       2645 &      0.124 &      75.74 \\

        10 &        100 &      10481 &     0.0687 &       3320 &      0.152 &      85.17 \\

        15 &        100 &       7925 &     0.4329 &       4918 &      0.205 &      70.21 \\

        20 &        100 &      10565 &      0.171 &       6543 &       0.27 &      97.91 \\

        25 &        100 &      13205 &     0.0215 &       8167 &      0.314 &      95.87 \\
\hline
         2 &        500 &       1061 &     0.0243 &        707 &      0.051 &      97.71 \\

         4 &        500 &       2117 &     0.1181 &       1348 &      0.074 &      80.01 \\

         8 &        500 &       4229 &     0.0728 &       2839 &      0.119 &      82.92 \\

        10 &        500 &       5285 &     0.2149 &       3291 &      0.142 &      91.72 \\

        15 &        500 &       7925 &     0.2473 &       4922 &      0.209 &      90.56 \\

        20 &        500 &      10565 &   0.00049 &       6559 &       0.25 &      99.06 \\

        25 &        500 &      13205 &     0.4568 &       8188 &      0.308 &      90.35 \\
\hline
         2 &       1000 &       1061 &     0.0719 &        707 &      0.079 &      88.86 \\

         4 &       1000 &       2117 &     0.0448 &       1348 &       0.12 &      83.11 \\

         8 &       1000 &       4229 &       0.11 &       2641 &      0.198 &         89 \\

        10 &       1000 &       5285 &     0.1236 &       3297 &       0.24 &      63.72 \\

        15 &       1000 &       7925 &     0.1985 &       4922 &      0.342 &      89.72 \\

        20 &       1000 &      10565 &     0.2887 &       6559 &      0.446 &      88.23 \\

        25 &       1000 &      13205 &     0.4402 &       8118 &       0.56 &       80.7 \\
\hline
\end{tabular}  
\end{center}
\end{table*}

\begin{figure}[htb!]
    \centering
    \includegraphics[width=0.45\textwidth]{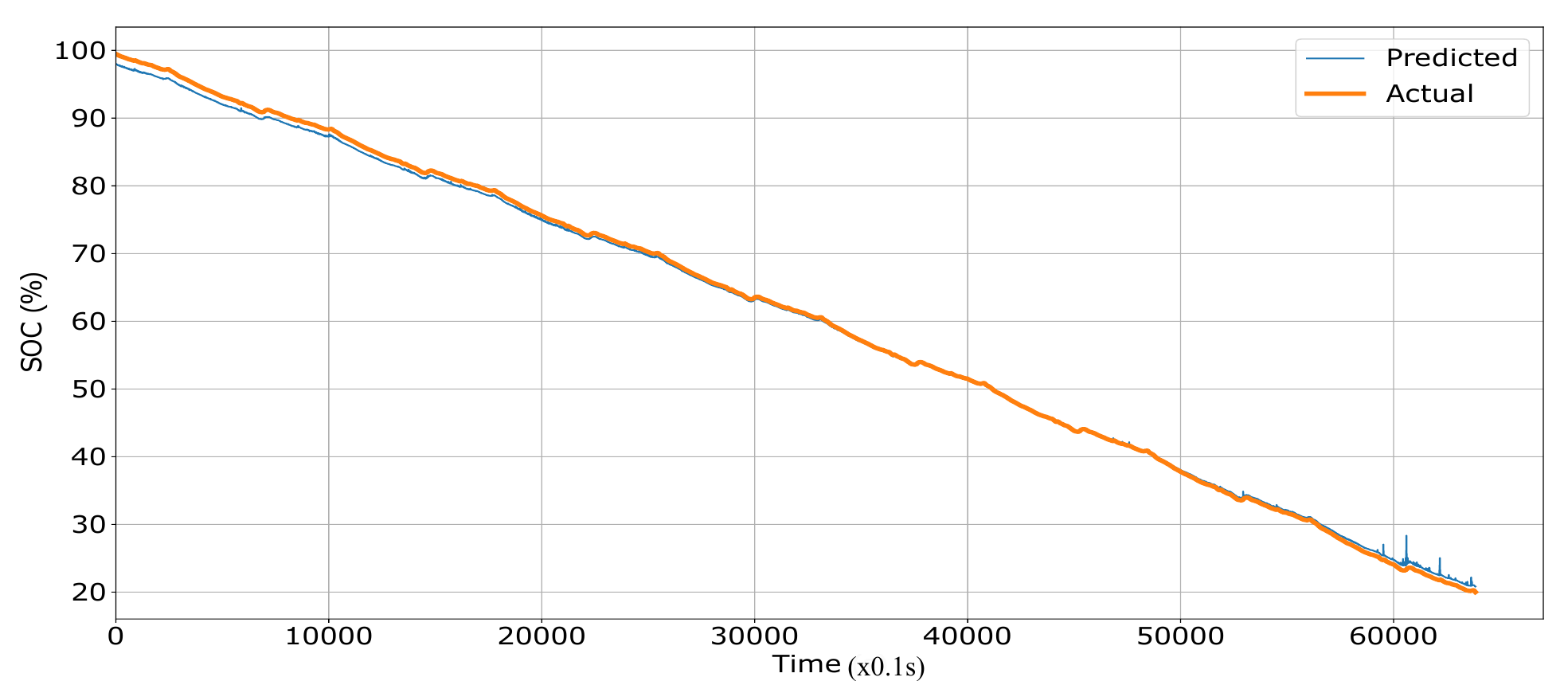}
    \caption{Prediction results on HWFET Drive Cycle}
    \label{fig:4}
\end{figure}

\begin{figure}[htb!]
    \centering
    \includegraphics[width=0.45\textwidth]{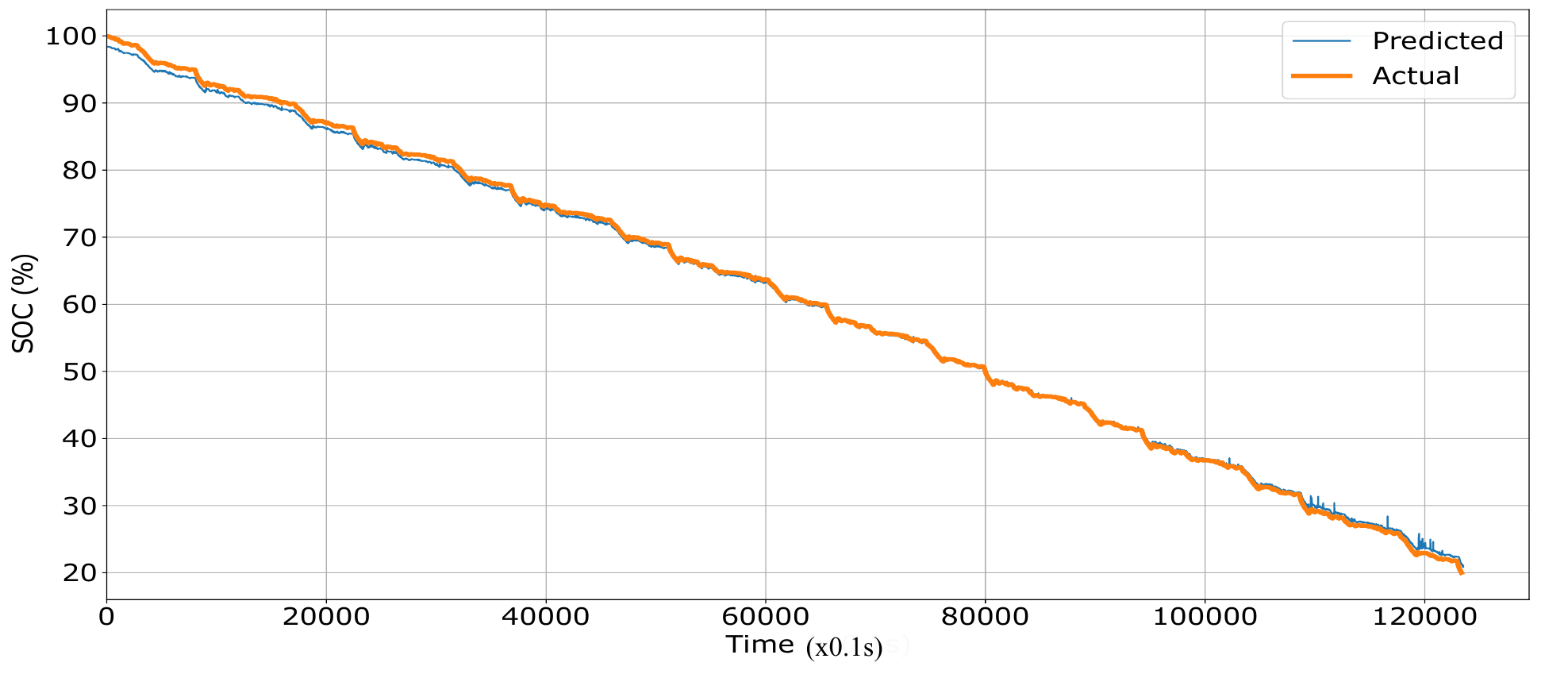}
    \caption{Prediction results on LA92 Drive Cycle}
    \label{fig:5}
\end{figure}

\begin{figure}[htb!]
    \centering
    \includegraphics[width=0.45\textwidth]{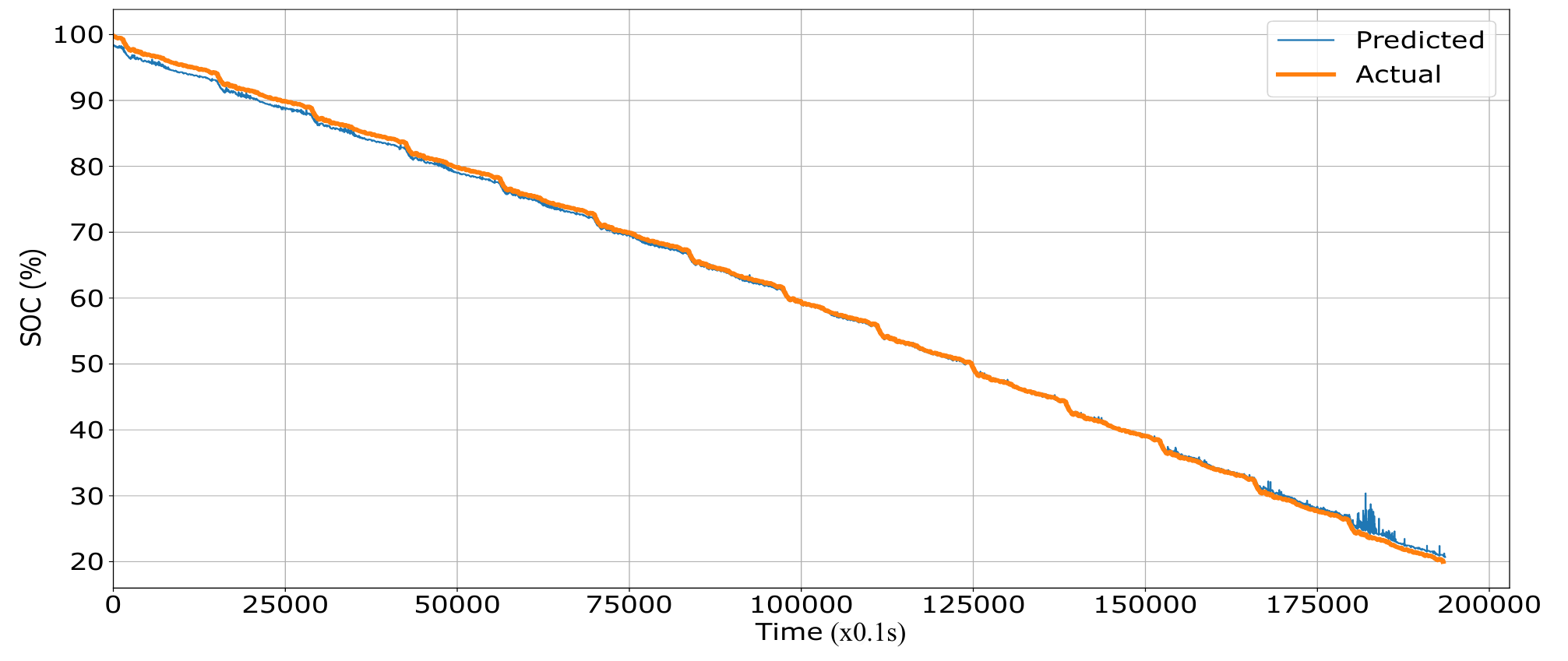}
    \caption{Prediction results on UDDS Drive Cycle}
    \label{fig:6}
\end{figure}

\begin{figure}[htb!]
    \centering
    \includegraphics[width=0.45\textwidth]{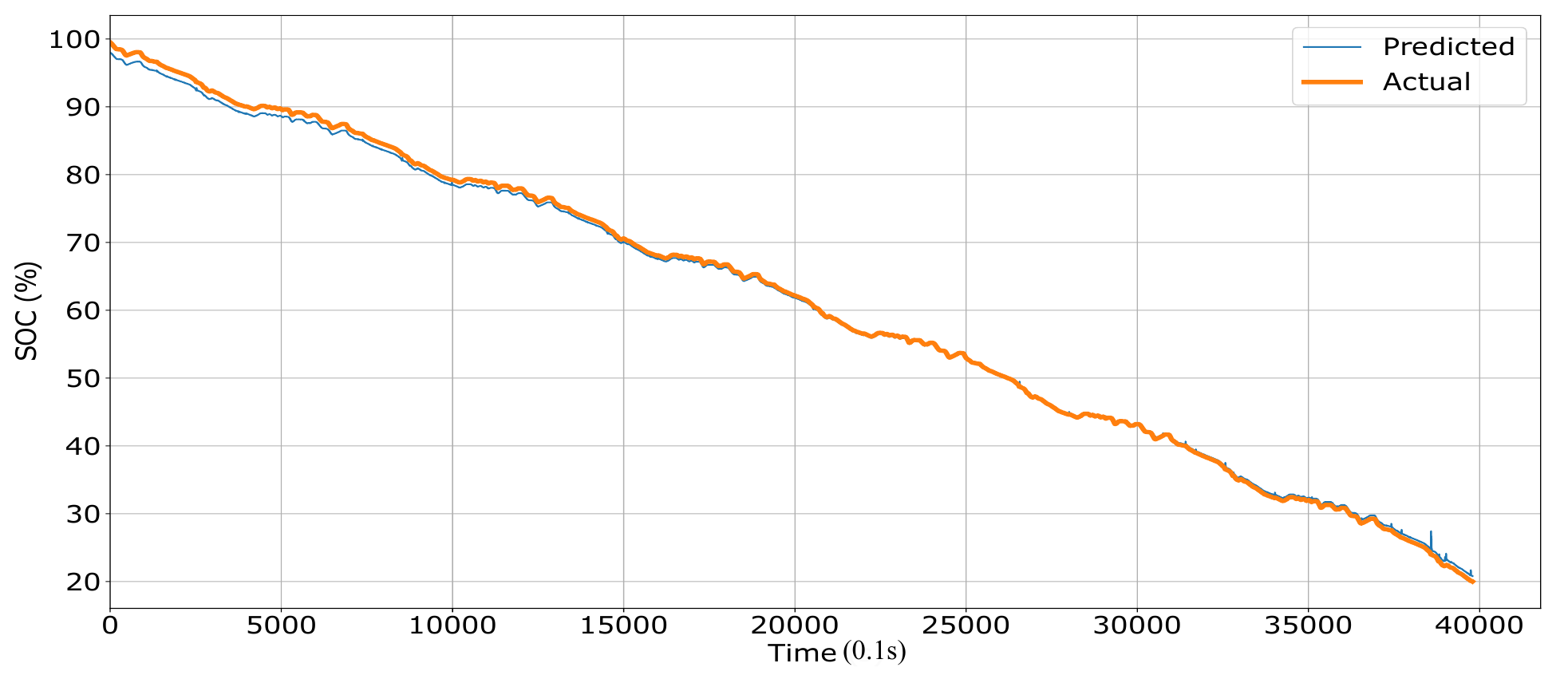}
    \caption{Prediction results on US06 Drive Cycle}
    \label{fig:7}
\end{figure}

\indent The results are shown in Table \ref{table:3}. The number of stacks is varied from 2 to 25 for the different input window lengths. The average accuracy for each model across the four drive cycles as well as the time taken by the model to make a single prediction is shown.\\
\indent The file size represents the memory space occupied by the trained weights and can be used to compare the relative memory space requirements between each model. As expected, increasing the number of stacks leads to a deeper network, which requires more memory space for storage. Also, the time taken to make a prediction increases, on account of the increased computational complexity of the model. However, even the deepest models with 25 stacks have a prediction time ranging from 0.31 ms to 0.56 ms, which is fast enough for real-time application on an EV.\\ 
\indent The fact that the models of the 500 input window length performed better on average than the 1000 input window length models is a surprising result. One would expect that increasing the look-back time would allow the model to capture more of the battery's dynamics. However, there is a noticeable decrease in accuracy at the 1000 time steps. The 500 input window length corresponds to about 50 seconds, which could belie the fact that the key battery dynamic to be modelled for time series prediction may occur within a time period of 50 seconds, and is subtle enough that it is missed at the higher input window of 100 seconds.\\ 
\indent Another counter-intuitive result was that the optimal number of stacks is 20 stacks, and not 25. It would be reasonable to assume that a deeper network would perform better, but the results show that the accuracy drops at this number of network layers. This could be attributed to insufficient data to train a network with 100 convolution layers, as the data becomes increasingly sparse at higher dimensions, or could be caused by over-fitting of the model.\\
\indent The best performing model was the model with 20 stacks and an input window of 500. The accuracy achieved was 99.1\%, representing a near perfect prediction fit. The graphs of this model’s performance on the various drive cycles is shown in Figs. \ref{fig:4} - \ref{fig:7}. The prediction error increases at lower SOC values, and this trend is shown in the figures. This can be attributed to the sparsity of data at lower SOC values, as only 10\% of the training data corresponded to SOC values less than 20\%. This is because of the unavailability of data below 12\% SOC. \\

\section{Conclusion}
\indent This work is the first to employ Temporal Convolution Networks for SOC estimation in EVs. The results show the feasibility of the TCN architecture for the SOC estimation task. This type of ML model also has the advantage of being computationally more efficient than other methods in the existing literature. The best performing model achieved an average accuracy of 99.1\% on the HWFET, LA92, UDDS and US06 drive cycles at 1 C and 25 \degree Celsius.\\
\indent Even-though TCN for SOC estimation has not received the amount of interest as other methods like NARXNN and LSTM-RNN have, the results are comparable for these models. This shows that several advancements in network structure and optimization could be implemented to further improve performance, and still be able to deploy the model on low-cost BMS hardware. Future work on this could also involve making the model robust to ambient temperature changes and variation in C rates.

\bibliographystyle{ieeetran}
\bibliography{references}

\end{document}